\documentclass[]{spie}  

\usepackage{amsmath,amsfonts,amssymb}
\usepackage{subcaption}
\usepackage{graphicx}
\usepackage{cite} 
\usepackage{epsfig}
\usepackage{amsmath}
\usepackage{nccmath}
\usepackage{amssymb}
\usepackage{mwe}
\usepackage{acro}
\usepackage{amssymb}
\usepackage{xcolor,colortbl}
\usepackage{tabularx}
\usepackage{relsize}
\usepackage{pifont}
\usepackage{booktabs} 
\usepackage{multirow}
\usepackage{multicol}
\usepackage{adjustbox}
\usepackage{float}
\usepackage{graphicx}
\usepackage{makecell}
\usepackage{tabu}
\usepackage[colorlinks=true, allcolors=blue]{hyperref}
\usepackage[capitalize]{cleveref}

\title{Optimizing Transmit Field Inhomogeneity of Parallel RF Transmit Design in 7T MRI using Deep Learning}
   
\author[a]{Zhengyi Lu}
\author[b,c]{Hao Liang}
\author[d]{Xiao Wang}
\author[b,c,a]{Xinqiang Yan}
\author[e,a]{Yuankai Huo}

\affil[a]{Department of Electrical and Computer Engineering, Vanderbilt University, Nashville, TN, USA}
\affil[b]{Vanderbilt University Institute of Imaging Science, Vanderbilt University Medical Center, Nashville, TN, USA}
\affil[c]{Department of Radiology and Radiological Sciences, Vanderbilt University Medical Center, Nashville, TN, USA}
\affil[d]{Computational Science and Engineering Division, Oak Ridge National Laboratory, Oak Ridge, TN, USA}
\affil[e]{Department of Computer Science, Vanderbilt University, Nashville, TN, USA}

\authorinfo{Corresponding author: Yuankai Huo: E-mail: yuankai.huo@vanderbilt.edu}

\begin{document} 
\maketitle

\begin{abstract}

Ultrahigh field (UHF) Magnetic Resonance Imaging (MRI) provides a higher signal-to-noise ratio and, thereby, higher spatial resolution. However, UHF MRI introduces challenges such as transmit radiofrequency (RF) field ($B_{1}^{+}$) inhomogeneities, leading to uneven flip angles and image intensity anomalies. These issues can significantly degrade imaging quality and its medical applications. This study addresses $B_{1}^{+}$ field homogeneity through a novel deep learning-based strategy. Traditional methods like Magnitude Least Squares (MLS) optimization have been effective but are time-consuming and dependent on the patient’s presence. Recent machine learning approaches, such as RF Shim Prediction by Iteratively Projected Ridge Regression and deep learning frameworks, have shown promise but face limitations like extensive training times and oversimplified architectures. We propose a two-step deep learning strategy. First, we obtain the desired reference RF shimming weights from multi-channel $B_{1}^{+}$ fields using random-initialized Adaptive Moment Estimation. Then, we employ Residual Networks (ResNets) to train a model that maps $B_{1}^{+}$ fields to target RF shimming outputs. Our approach does not rely on pre-calculated reference optimizations for the testing process and efficiently learns residual functions. Comparative studies with traditional MLS optimization demonstrate our method's advantages in terms of speed and accuracy. The proposed strategy achieves a faster and more efficient RF shimming design, significantly improving imaging quality at UHF. This advancement holds potential for broader applications in medical imaging and diagnostics.

\end{abstract}


\section{INTRODUCTION}
\label{sec:intro}  
In Magnetic Resonance Imaging (MRI), Parallel Transmission (PTx) is a key technology used to generate high-quality images at ultra high fields (UHF)~\cite{Zhu2004_add, Katscher2006_add}. As the strength of the static field increases, the wavelength of the RF field becomes comparable to the imaged human tissues, which causes destructive interferences~\cite{VandeMoortele2005_add}. Such destructive interferences lead to an inhomogeneous $B_{1}^{+}$ field and, thereby, uneven flip angles and anomalous image intensity. If left unsolved, the non-uniformity will decrease the quality of imaging at ultra high fields, thus affecting its application in medical fields.

$B_{1}^{+}$ field homogeneity can be adequately achieved with an n-element coil array by RF shimming on single slices or over the whole volume~\cite{Mao2006}. In cases that focus solely on the image magnitudes, only $B_{1}^{+}$ fields of each coil for each subject need to be measured to adjust the phases and amplitudes for each scan geometry~\cite{4Setsompop2008}. To solve for $B_{1}^{+}$ field uniformity, Setsompop et al. introduced a magnitude least squares (MLS) optimization method for parallel RF excitation at 7 Tesla using an eight-channel transmit array, which provides an enhanced magnitude profile~\cite{4Setsompop2008}. The MLS problem in RF pulse design parallels the phase retrieval problem, which is identified as a useful technique in communities~\cite{4Setsompop2008, 5Guerin2014, 6Kerr2007, 7Cloos2012}. Based on the MLS optimization, various further types of research have been established to solve for uniform magnetization magnitude~\cite{9Grissom2012, 10Padormo2016}. However, such a kind of optimization needs to be done upon the patient’s presence on the scanner and takes much time for target RF shimming weights calculation~\cite{8Cao2016, 11Kilic2024}.

In recent years, machine learning-based methods have been applied to solve the inhomogeneity problem from the scanned $B_{1}^{+}$ field to gain target magnetization~\cite{20Mirfin2018, 21Shin2021, Ianni2018}. One such kind of approach called RF Shim Prediction by Iterative Projected Ridge Regression (PIPRR) was proposed by Ianni et al., which integrates the design of training shims with learning to enable interpolation over training shims~\cite{Ianni2018}. However, this approach compromises training time to seek fast prediction per slice, which took 5 days without parallelization across training slices. Another work proposed a deep learning (DL) framework that predicts $B_{1}^{+}$ distributions after within-slice motion, which enables real-time tailored pulse redesign and significantly reduces motion-related excitation errors~\cite{12Plumley2022}. An unsupervised deep learning method applied a convolutional neural network (CNN) architecture that minimizes the discrepancy between the Bloch simulation output and the target magnetization with a physics-driven loss function~\cite{12Plumley2022}. However, it does not take the input power, voltage, and specific absorption rate (SAR) into consideration, and the total architecture is relatively simple.

In this work, we designed a deep learning-based two-step strategy which first gains the desired reference RF pulse from multi-channel $B_{1}^{+}$ magnetic field over slices with random-initialized Adaptive Moment Estimation~\cite{17Kingma2014}, then applies Residual Networks (ResNets) proposed by He et al. to train a model which takes $B_{1}^{+}$ field as input and target RF shimming as output~\cite{13He2016}. A comparison study is also done to monitor the Root Mean Square Error (RMSE) and operating time between our strategy and the traditional MLS optimization method along slices. The contribution of this paper is twofold:

\begin{itemize}
\item A random-initialized Adaptive Moment Estimation is applied to find reference RF shimming weights, which avoids local minima that often occur in the MLS algorithm~\cite{17Kingma2014}.

\item A ResNet architecture~\cite{13He2016} that enables the network to learn residual functions from the $B_{1}^{+}$ magnetic field to the $B_{1}^{+}$ complex RF shimming weights for each channel.

\end{itemize}

\begin{figure*}[t]
\begin{center}
\includegraphics[width=0.7\linewidth]{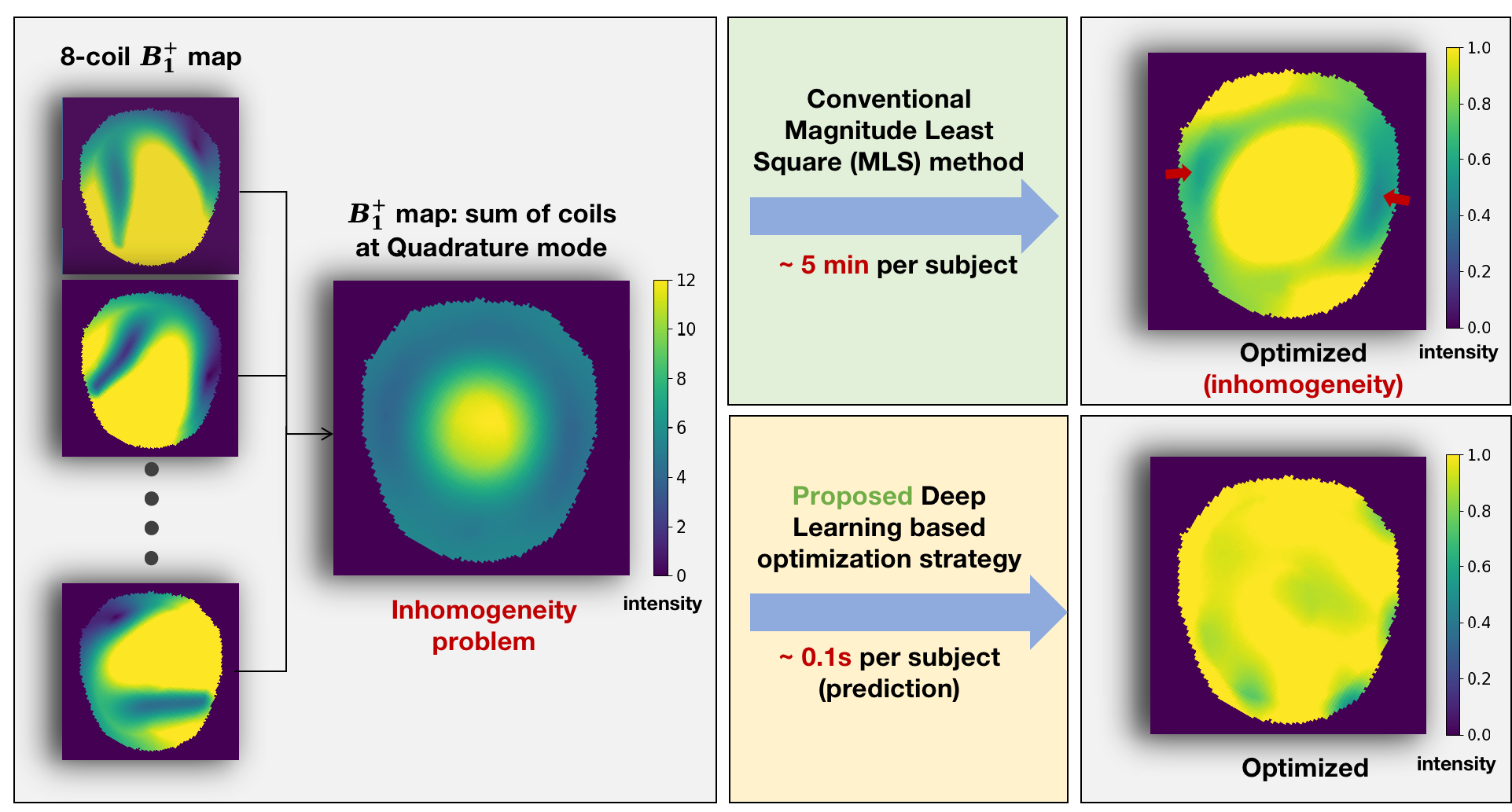}
\end{center}
\caption{The $B_{1}^{+}$ field inhomogeneity problem addressed in RF shimming design and main advantages of our proposed strategy over conventional MLS method. The image to the right presents the 8-coil $B_{1}^{+}$ data we started with. The two central blocks show our method and conventional MLS method with costed runtime over one subject which is estimated from testing prediction. On the right are two optimization results on the same slice picked from testing cases, which shows a contrast of uniformity over the $B_{1}^{+}$ field between the two methods.}
\label{fig:problem}
\end{figure*}

\section{METHOD}
\label{sec:method}  

\subsection{Magnitude least squares optimization}
MLS optimization is a specific approach used in RF shimming to minimize the difference between the desired and actual magnitudes of the $B_{1}^{+}$ field across the imaging volume~\cite{4Setsompop2008}. Unlike other methods that might focus on the phase or real/imaginary components~\cite{15LeRoux1998, 16Collins2005}, MLS specifically targets the magnitude of the field. The $B_{1}^{+}$ field produced by each coil element is modeled, typically using the Biot-Savart law or electromagnetic simulations. The total $B_{1}^{+}$ field at any point in the imaging volume is the vector sum of the fields produced by all the coils. After the modeling, the optimization can be expressed as~\cite{4Setsompop2008, 14Zheng2012} 

\begin{equation}\label{b}
\begin{split}
    b(t) = \arg\min_b \left\{ \left\| \left| Ab \right| - m \right\|_w^2 + \lambda \left\| b \right\|^2 \right\}
\end{split}
\end{equation}

\noindent where $A$ is the matrix that contains the whole $B_{1}^{+}$ field values at each spatial location for each coil, $b$ is the weights for each coil in RF shimming design we want to solve for, $m$ is the target magnetic field map, $w$ is a mask used to specify the region of the brain from surroundings, and $\lambda$ is the regularization parameter used to balance the RF power and excitation errors~\cite{14Zheng2012}.

\begin{figure*}[t]
\begin{center}
\includegraphics[width=0.9\linewidth]{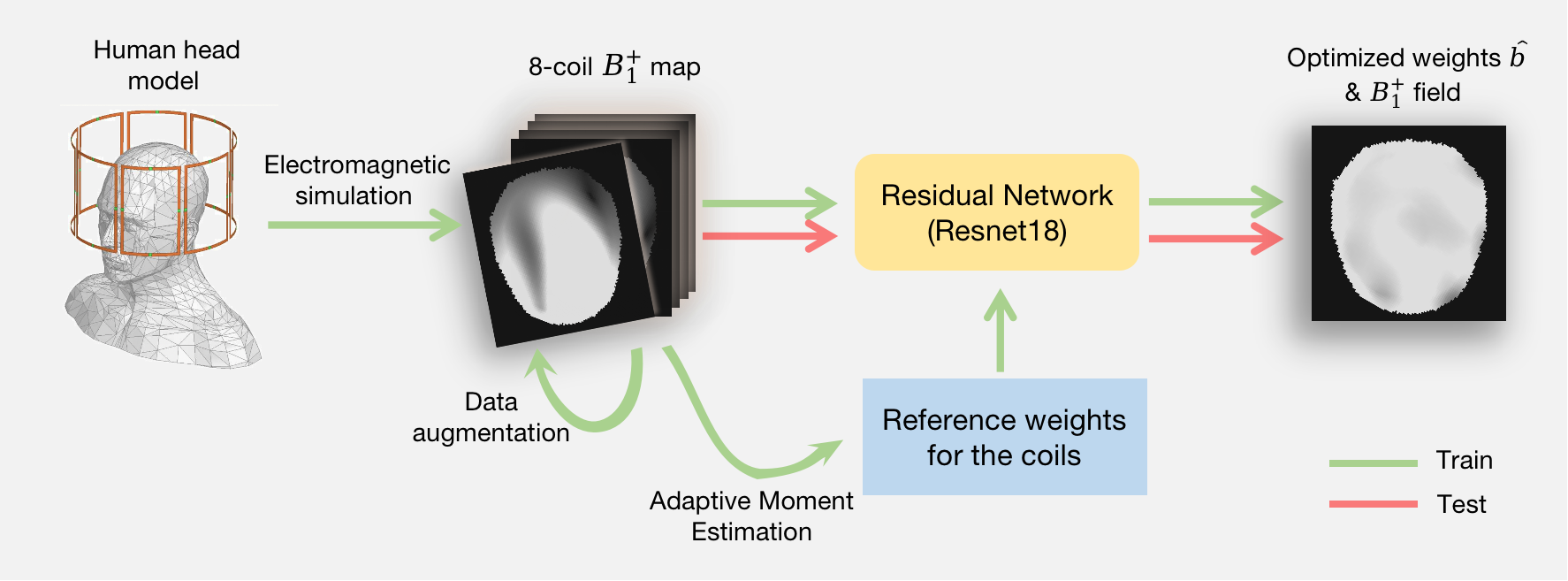}
\end{center}
\caption{The procedures of training and prediction of our proposed optimization strategy. The strategy starts with simulations to get the desired $B_{1}^{+}$ as inputs of the Residual Network (ResNet18). Data augmentation is then applied. Adaptive Moment Estimation~\cite{17Kingma2014} is used to calculate the reference weights of coils as targets in training. At last, predictions are made by applying the testing data into the trained DL model.}
\label{fig:problem}
\end{figure*}

\subsection{Quadrature mode and random initialization}
The optimization process of random-initialized Adaptive Moment Estimation and MLS algorithm for compression are both started with the coil elements arranged in a quadrature configuration. The quadrature configuration is obtained according to the setup of coils during the simulation in the data generation process as shown in \ref{fig:problem}. The 8 RF coils are positioned at 45 degrees to each other around the scanning subject~\cite{19Ibrahim2005}.

To increase the chance of finding the best minimum, we utilize random initialization for Adaptive Moment Estimation~\cite{17Kingma2014} when generating reference RF shimming weights. Three hundred random vectors representing weights for coils are generated, which aims to escape some local minimum.

\section{EXPERIMENTS}
\label{sec:experiment}

\subsection{Data preparation}
Electromagnetic simulations were performed using a commercial FEM-based Maxwell solver (Ansys HFSS, Canonsburg, PA, USA). The transmit array consisted of 8 loop elements configured in a single-row layout. All transmit coils were mounted on a 28 cm diameter cylinder to accommodate a receive coil, with each element measuring 16×10 cm². The coils were spaced approximately 1 cm apart. The standard human body model from Ansys was used, with its dimensions scaled in 10 scenarios. The models represented average male and female dimensions from five countries. The transmit array was modeled at 298 MHz, corresponding to the Larmor frequency for 7T. The voxel size within each object was ensured to be the same among the 10 objects. After all simulations, 10 8-channel $B_{1}^{+}$ magnetic field volumes were obtained together with slices, each volume is 101×101×71×8 in size. The mass density of 10 head models is also provided.

With the $B_{1}^{+}$ fields, we performed several procedures for data pre-processing before applying the algorithm. For each head object, we went over the 71 slices and selected 32 good cases to eliminate errors during simulation. Then binary masks were generated according to the mass density map of each subject to preserve the desired regions (skull, brain, etc.) and exclude the air outside of the head. Data augmentation like rotation was used to generate more data for the project. Finally, 3860 masked $B_{1}^{+}$ map slices were generated for further calculation.

The $B_{1}^{+}$ fields of each channel were set according to quadrature mode. For each fold of experiments, to obtain the weights used for training, adaptive moment estimation was used to minimize the loss defined in Eq.~\ref{b}. The optimization was repeated with random initial weights, and the weights yielding the best results were selected as the training target.
For comparison, a single MLS optimization for each fold was performed for each sample. The weights calculated from MLS were used to perform numerical evaluation as the contrast experiment.

\subsection{Networks, training, and testing}
Resnet18~\cite{13He2016} network architectures were used to predict the RF coil weights from $B_{1}^{+}$ maps by learning the transformation between the input data and target weights. An input size of 101×101×32 and an output size were used in the architectures. It begins with an initial convolutional layer, followed by four stages of residual blocks.

For each stage, it has two BasicBlocks, with each block containing two 3×3 convolutional layers, batch normalization, and ReLU activation. The feature map sizes are 64, 128, 256, and 512, doubling each stage. Downsampling is done using a stride of two in the first block of each stage, with 1×1 convolutions for dimension matching. The final layers include adaptive average pooling to 1×1, followed by a fully connected layer to the 32 output classes, ensuring efficient gradient flow and training.

During training, the loss function is defined as the mean square error between the predicted and reference RMSE:

\begin{equation}\label{RMSE}
\begin{split}
\text{RMSE} = \sqrt{ \frac{ \left\| \left| \mathbf{A} \mathbf{b} \right| - \mathbf{m} \right\|_{\mathbf{w}}^2 }{ N_{\text{voxel}} } }
\end{split}
\end{equation}

\begin{equation}\label{loss}
\begin{split}
\text{loss} = \frac{1}{N_{\text{slice}}} \sum_{i=1}^{N_{\text{slice}}} \left| {\text{RMSE}_{\text{pred}}}^{(i)} - {\text{RMSE}_{\text{ref}}}^{(i)} \right|
\end{split}
\end{equation}

\noindent where RMSE is calculated from Eq.~\ref{b}, $N_{\text{voxel}}$ is the number of voxels in the current slice, ${\text{RMSE}_{\text{pred}}}^{(i)}$ is the RMSE calculated from predicted weights on the $i_{th}$ slice, ${\text{RMSE}_{\text{ref}}}^{(i)}$ is the RMSE calculated from reference weights, and $N_{\text{slice}}$ is the number of slices.

The entire dataset was randomly separated as 8:1:1 for training, validation, and testing with a batch size of 16. The separation and experiment process (training and testing) were done five times to avoid accidental cases. Adaptive Moment Estimation~\cite{17Kingma2014} optimizer was used in our training with an initial learning rate of $10^{-3}$ and a decay rate of $50\%$ every 50 epochs for a total of 200 epochs. All training  and testing were performed with PyTorch on an NVIDIA GeForce RTX A6000 (Cuda 12.3).

\begin{table}[h]
\centering
\begin{tabular}{c|cc|cc|cc|cc}
\hline\hline
 & \multicolumn{2}{c|}{Mean RMSE} & \multicolumn{2}{c|}{Best RMSE} & \multicolumn{2}{c|}{Worst RMSE} & \multicolumn{2}{c}{Runtime(per slice)} \\
\cline{2-9}
Fold & MLS & Proposed & MLS & Proposed & MLS & Proposed & MLS & Proposed \\
\hline
1 & 10.4082 & 9.4263 & 7.1944 & 6.8553 & 15.6868 & 14.9016 & 4.1146 (s) & 1.4237 (ms) \\
2 & 10.1321 & 9.1191 & 7.2148 & 6.7598 & 16.4810 & 14.9022 & 4.3769 (s) & 1.4516 (ms) \\
3 & 10.1343 & 9.1823 & 7.1944 & 6.7918 & 16.1520 & 14.5706 & 3.7555 (s) & 1.4384 (ms) \\
4 & 10.2196 & 9.0711 & 7.1944 & 6.7910 & 15.6781 & 14.9342 & 3.8146 (s) & 1.4533 (ms) \\
5 & 10.2289 & 8.9203 & 7.2353 & 6.6331 & 16.1629 & 13.6948 & 3.6996 (s) & 1.4442 (ms) \\
\hline\hline
\end{tabular}
\caption{Comparison of metrics between MLS and proposed methods. In each trail, the MLS method and proposed method are tested on the same dataset with the same separation. The Mean RMSE, Best RMSE and Worst RMSE are all calculated corresponding to the target, measured by [\% of Target FA], where a lower RMSE means a better performance in magnetic field homogeneity. The runtime are recorded under the same tasks and then calculated per slice. A statistical test was conducted, resulting in a significant difference with $p < 0.001$.}
\label{tab:table}
\end{table}

\begin{figure*}[t]
\begin{center}
\includegraphics[width=0.9\linewidth]{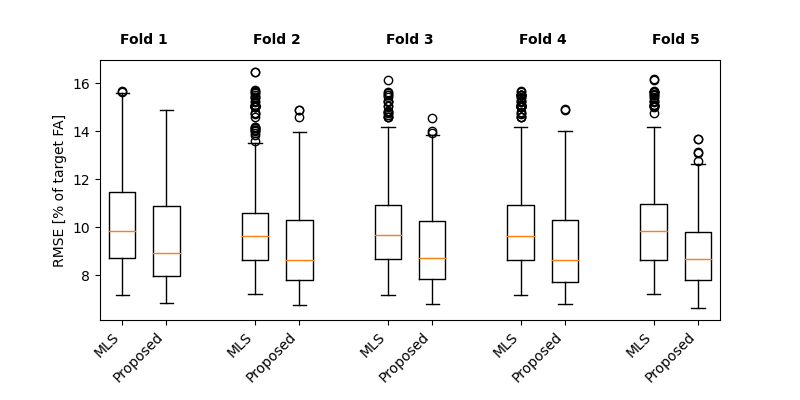}
\end{center}
\caption{The RMSE results [\% of target FA] of MLS method and proposed method are shown in the box-plots. Five folds of testing results are drawn independently in five groups in the figure. A significant difference with $p < 0.001$ is ensured.}
\label{fig:boxplot}
\end{figure*}

\section{RESULTS}
The proposed deep learning-based optimization strategy was evaluated against the traditional MLS method across five folds with different separation, focusing on Mean RMSE, Best RMSE, Worst RMSE over the testing dataset slices, and runtime in Tab.~\ref{tab:table}.

The proposed method consistently achieved lower Mean RMSE values, ranging from 8.9203 to 9.4263 [\% of target FA], compared to the MLS method's range of 10.1321 to 10.4082. Similarly, the Best RMSE for the proposed method ranged from 6.6331 to 6.8553 [\% of target FA] outperforming the MLS method's range of 7.1944 to 7.2353 [\% of target FA] In terms of Worst RMSE, the proposed method exhibited values between 13.6948 and 14.9342 [\% of target FA], while the MLS method showed higher values, ranging from 15.6686 to 16.4810 [\% of target FA]. The runtime for the proposed method was significantly reduced, averaging around 1.44 milliseconds per slice, compared to the MLS method's around 4 seconds per slice. The total runtime for a scanned subject (71 slices) can be estimated according the average runtime in testing cases.

As shown in Fig.~\ref{fig:boxplot}, the RMSE results in format of box plots further supports these findings. The proposed method consistently showed lower median RMSE values and tighter interquartile ranges than the MLS method, underscoring its effectiveness.

From the above results, we proved that the proposed method not only reduces RMSE significantly but also achieves much faster computation times compared to the MLS method, offering a more efficient and accurate solution for RF shimming in ultra-high field MRI.

\label{sec:result}

\section{CONCLUSION}
The study shows that deep-learning networks trained with $B_{1}^{+}$ field data are capable of estimating the RF weights of coils in multi-coil RF shimming design to solve the inhomogeneity problem. Our proposed optimization strategy with deep learning network outperforms the conventional method of MLS method with a quadrature initialization~\cite{4Setsompop2008}. By applying our method on the simulated $B_{1}^{+}$ filed data, we showed that our method requires less inference time, while maintaining a lower RMSE value, compared with traditional MLS optimization.

\label{sec:conclusion}

\acknowledgments 
This research was supported by NIH R01DK135597(Huo), DoD HT9425-23-1-0003(HCY), NIH NIDDK DK56942(ABF). This work was also supported by Vanderbilt Seed Success Grant, Vanderbilt Discovery Grant, and VISE Seed Grant. This project was supported by The Leona M. and Harry B. Helmsley Charitable Trust grant G-1903-03793 and G-2103-05128. This research was also supported by NIH grants R01EB033385, R01DK132338, REB017230, R01MH125931, and NSF 2040462. We extend gratitude to NVIDIA for their support by means of the NVIDIA hardware grant. This works was also supported by NSF NAIRR Pilot Award NAIRR240055. This manuscript has been co-authored by ORNL, operated by UT-Battelle, LLC under Contract No. DE-AC05-00OR22725 with the U.S.Department of Energy.
This paper describes objective technical results and analysis. Any subjective views or opinions that might be expressed in the paper do not necessarily represent the views of the U.S. Department of Energy or the United States Government. 
\bibliography{report} 
\bibliographystyle{spiebib} 

\end{document}